%% file: main.tex
\documentclass[10pt,twocolumn,letterpaper]{article}

\usepackage[pagenumbers]{cvpr} 
\usepackage[accsupp]{axessibility}

\input{preamble}

\definecolor{cvprblue}{rgb}{0.21,0.49,0.74}
\usepackage[pagebackref,breaklinks,colorlinks,citecolor=cvprblue]{hyperref}

\def\paperID{10924} 
\def\confName{CVPR}
\def\confYear{2024}

\title{DSGG: Dense Relation Transformer for an End-to-end Scene Graph Generation}

\author{Zeeshan Hayder$^{1}$,
Xuming He$^{2}$, \\
$^{1}$Data61-CSIRO, Australia, $^{2}$ShanghaiTech University, \\
{\tt\small zeeshan.hayder@data61.csiro.au}, {\tt\small hexm@shanghaitech.edu.cn}
}

\begin{document}

\maketitle

\begin{abstract}
Scene graph generation aims to capture detailed spatial and semantic relationships between objects in an image, which is challenging due to incomplete labelling, long-tailed relationship categories, and relational semantic overlap. Existing Transformer-based methods either employ distinct queries for objects and predicates or utilize holistic queries for relation triplets and hence often suffer from limited capacity in learning low-frequency relationships. In this paper, we present a new Transformer-based method, called DSGG, that views scene graph detection as a direct graph prediction problem based on a unique set of graph-aware queries. In particular, each graph-aware query encodes a compact representation of both the node and all of its relations in the graph, acquired through the utilization of a relaxed sub-graph matching during the training process. Moreover, to address the problem of relational semantic overlap, we utilize a strategy for relation distillation, aiming to efficiently learn multiple instances of semantic relationships. Extensive experiments on the VG and the PSG datasets show that our model achieves state-of-the-art results, showing a significant improvement of 3.5\% and 6.7\% in mR@50 and mR@100 for the scene-graph generation task and achieves an even more substantial improvement of 8.5\% and 10.3\% in mR@50 and mR@100 for the panoptic scene graph generation task. Code is available at \url{https://github.com/zeeshanhayder/DSGG}.
\end{abstract}

\section{Introduction}

\begin{figure}
  \centering
  \begin{tabular}{c}
     \includegraphics[width=\linewidth, trim=5 27 0 0, clip]{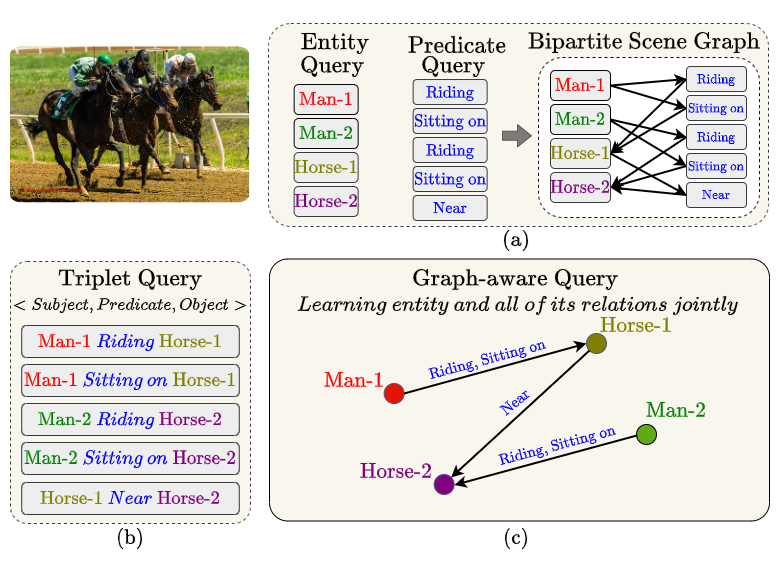}
  \end{tabular}
  \vspace{-3mm}
  \caption{Illustration of different queries used in SGG networks. a) Multi-query transformer networks learn entities and predicates separately.  
  b) Triplet query-based transformer networks use a separate query for each triplet. c) Our proposed graph-aware queries learn a compact representation of objects and all of its relations jointly.}
  \vspace{-7mm}
  \label{fig:graphquery}
\end{figure}

Scene Graph Generation (SGG)~\cite{tang2020sggcode} seeks to detect and generate a graph-structured summary of all the objects present in the scene, with edges describing their visual interactions or pairwise relationships. This topological representation of an image is helpful for visual understanding and image reasoning tasks such as image caption generation, visual question answering, cross-model retrieval, and human-object interaction recognition. This task is analogous to (or a sub-task of) the panoptic scene graph generation (PSG)~\cite{yang2022panoptic} task, where the subjects and objects can also belong to the stuff classes and the semantic segmentation of the entities are used to evaluate the scene graphs.  

Specifically, given an image, the SGG task focuses on predicting all the objects along with their class labels, bounding boxes, pixel-accurate segmentation, and their relations to all other objects. The graph-based scene generation methods~\cite{zellers2018scenegraphs, tang2018learning} are often limited by the high complexity of the underlying object detectors~\cite{ren2016faster} and by the representation of the scene context. Unbiased SGG methods~\cite{tang2020unbiased} attempt to learn the semantic relationships without considering the label bias in the data and then use simple post-processing to correct the label distribution. Nevertheless, these techniques have challenges with images that have multiple semantic relations between the same pair of objects and are vulnerable to long-tail problems with relation categories. Transformer-based methods~\cite{zellers2018scenegraphs, tang2020unbiased, Li_2021_CVPR, dong2022stacked, Li_2022_CVPR, Li0HZZ022, lin2022hl, yang2022panoptic, zhou2023hilo} attempt to provide a single-stage solution for scene graph generation. Traditional transformer-based approaches utilize a two-stream network with either a shared query or separate query for estimating the object relations~\cite{zellers2018scenegraphs, tang2020unbiased, Li_2021_CVPR, dong2022stacked}. Recent approaches, such as~\cite{Teng_2022_CVPR, lin2022hl, lin2022ru, HetSGG, Li0HZZ022, yang2022panoptic, Li_2022_CVPR, zheng2023prototype}, adopt a holistic strategy by directly predicting a list of <subject, predicate, object> triplets, with each query in the network representing a singular triplet. Recent transformer-based scene graph generation approaches, ~\cite{Relationformer, liu2023repsgg}, depend solely on object-based matching to learn queries. However, their constrained capacity results in limitations on effectively learning dense and low-frequency relations.

HiLo~\cite{zhou2023hilo} introduced a two-stream network based on triplet queries that use an ad-hoc approach to add pseudo-relations to address the relation class imbalance. Nevertheless, it fails to comprehensively capture all relations in the image and remains vulnerable to the issue of relational semantic overlap, even when employing dedicated network branches for low-to-high and high-to-low frequency relations. Another limitation is the model's capacity to capture the substantial diversity within each relation category and the similarities in relations that exist across multiple objects.

In this paper, we address this gap by introducing a universal model that learns all relations among the image objects. Specifically, we introduce a graph-aware query, depicted in Figure~\ref{fig:graphquery}, which serves as a compositional query. This query learns the representation of each object along with its multiple relations to all other objects in the image. Essentially, each node in the graph has a unique graph-aware query associated to it. This is in contrast to the existing transformer-based architectures—whether they use single or triplet queries—that have difficulty scaling up to generate dense scene graphs because of the model's rising complexity with the traditional queries needed for every possible triplet. Using these graph-aware queries has the advantage that the models learn to predict the right multiple relation labels (or no relation), essentially eliminating the relational semantic overlap problem, regardless of whether multiple relations exist between two objects. An additional benefit is that overall number of trainable network parameters is reduced because nodes and relations do not require two-stream transformer.   

Moreover, learning these graph-aware queries in an end-to-end context is challenging. In this paper, we propose to expand the set prediction problem~\cite{carion2020end} to graph prediction based on graph-aware queries, which are essential for learning the scene graph's structure. To match every node and all of its edges in the learned graph with the ground truth node representation, a relaxed sub-graph matching technique is employed. In the presence of low-frequency relations, the sub-graph matching places greater emphasis on learning the overall graph structure than on the specific high-frequency relations present in the image, thus eliminating the long-tail relation distribution problem. In addition, the DSGG approach adapts a re-scoring mechanism and introduces relation distillation for effective pairwise relation prediction. As the model becomes more adept at filtering out negative relations, the label noise decreases by learning of the dense image relations across all objects in the image.

In summary, we propose DSGG, an end-to-end unified technique that investigates scene graph detection as a direct graph prediction problem and estimates multi-label relation probability for each pair of nodes in the graph. The main contributions of our work are fourfold:  
\begin{itemize}
    \item We introduce graph-aware queries for the transformer-based network that learns a compact representation of both the node and all of its relations in the graph. 
    \item A novel sub-graph matching is introduced to estimate the cost between ground truth and the estimated scene graph.
    \item A relation distillation is introduced and the re-scoring module is adapted to effectively filter and rank the predicates based on entities semantics. 
    \item With state-of-the-art performance on the Visual Genome and PSG datasets, our method considerably improves the visual semantic relations for both the scene graph detection and panoptic scene-graph generation tasks. 
\end{itemize}

\section{Related Work}

\subsection{Scene graph generation}
Most prior works on SGG are primarily focused on the Visual Genome (VG) dataset~\cite{XuZCF17} and use the benchmark suite~\cite{tang2020sggcode}. These methods can be broadly categorized into bottom-up, top-down, and hybrid approaches. 

\paragraph{Bottom-up Graph-based SGG approaches}
create the scene graph using a multi-stage process and mostly rely on an object detector. 
Traditional methods ~\cite{XuZCF17, zellers2018scenegraphs, tang2018learning, Liu2021FullyCS, pix2graph} organize object proposals generated by a detector into graph nodes. These nodes are further processed to learn the edge context and predict pairwise relationships. The high complexity of the traditional object detectors~\cite{ren2016faster} and redundant object proposals make these approaches relatively complex and inefficient. Modern graph-based approaches~\cite{lin2022ru, HetSGG} attempt to use a relaxed message-passing algorithm either in a full graph or a bipartite graph to infer the relation-aware context~\cite{lin2022ru}. 
Furthermore,~\cite{HetSGG} suggests aggregating the global contextual information of an image while taking the predicate type between objects into consideration. 

\paragraph{Bottom-up Query-based SGG approaches}
are based on detection transformer~\cite{Li_2022_CVPR, Li0HZZ022, lin2022hl, RelTr}, 
and are usually based on a shared encoder and train a separate decoder for object and relation heads. Specifically, the relation head learns a fixed set of queries.
These methods approach the object and the relation feature learning as independent branches~\cite{RelTr} and are limited by the explicit modeling of the pairwise relationships. Furthermore, ~\cite{CoRFT, liu2023repsgg, Relationformer} are dependent on predicted objects before relation classification among them. Specifically,~\cite{Relationformer} learns an extra [rln]-token in addition to [obj]-tokens, and ~\cite{liu2023repsgg} learns $K$ distinct queries for $K$ triplets for each object. 

\paragraph{Top-down SGG approaches}
~\cite{Desai2021LearningOV, desai2022singlestage, Li_2021_CVPR, Teng_2022_CVPR} relies on predicting triplet queries by detecting the relation proposals directly using a pre-trained object detector. 
In particular,~\cite{Desai2021LearningOV} focuses on low-frequency relations.~\cite{Li_2021_CVPR} generates the entity and predicate proposals separately, resulting in a bipartite graph neural network. A two-pass confidence gating technique is used, propagating messages from entities to predicates and back to entities, to learn the relations.
~\cite{Teng_2022_CVPR} combines global context information into objects via object pair fusion and an entity-to-relation fusion module. 
However, due to the highly coupled nature, these approaches have a limited potential to scale to dense scene graph generation.

\paragraph{Hybrid SGG approaches}, such as~\cite{Teng_2022_CVPR, dong2022stacked, zheng2023prototype, zhou2023hilo} mainly focus on two branch networks, that predict the object queries as well as triplet queries in parallel. These methods often require an ad-hoc approach to match the queries predicted in both branches.~\cite{Teng_2022_CVPR} adopt a bottom-up object detection network as an aid in a Siamese network for knowledge distillation.~\cite{dong2022stacked} proposed the use of multiple modalities (including language and image features) to learn both intermodal interaction and intramodal refinement.~\cite{zheng2023prototype} learns a common embedding space between subject-object pairs and predicates for relation detection.

\subsection{Panoptic scene graph generation}
The panoptic scene graph generation (PSG) task incorporates semantic segmentation for evaluation and includes stuff classes to the objects of interest, extending the SGG task. Note that, despite the similarity between the SGG and the PSG tasks, few strategies~\cite{zellers2018scenegraphs, XuZCF17, tang2020unbiased, yang2022panoptic, zhou2023hilo}  have been demonstrated to be effective for both tasks. We classify them into the following categories in general. 

\paragraph{Traditional approaches} extends the existing SGG methods with additional segmentation branch. ~\cite{yang2022panoptic} provides a comprehensive benchmark of two-stage~\cite{zellers2018scenegraphs, XuZCF17, tang2020unbiased, lin2020gps} and one-stage~\cite{yang2022panoptic} baseline methods. The segmentation branch is typically fine-tuned using a pre-trained MaskFormer~\cite{cheng2021maskformer}. 

\paragraph{Pseudo-label based approach}
as introduced in~\cite{zhou2023hilo}, utilize high-low and low-high pseudo-labels to learn diverse relation triplets. In contrast to the traditional segmentation model~\cite{cheng2021maskformer}, HiLo uses Mask2Former~\cite{cheng2021mask2former} segmentation model. In this paper, we also initialized DSGG model with the Mask2Former~\cite{cheng2021mask2former} segmentation branch for this task. 

In summary, our proposed method is a bottom-up graph-aware query-based model for scene graph generation that takes a direct-graph prediction approach without using any additional pseudo-relation labels for the training.

\section{Method}
\label{sec:pre}

\begin{figure*}
  \centering
  \begin{tabular}{c}
     \includegraphics[width=0.9\linewidth]{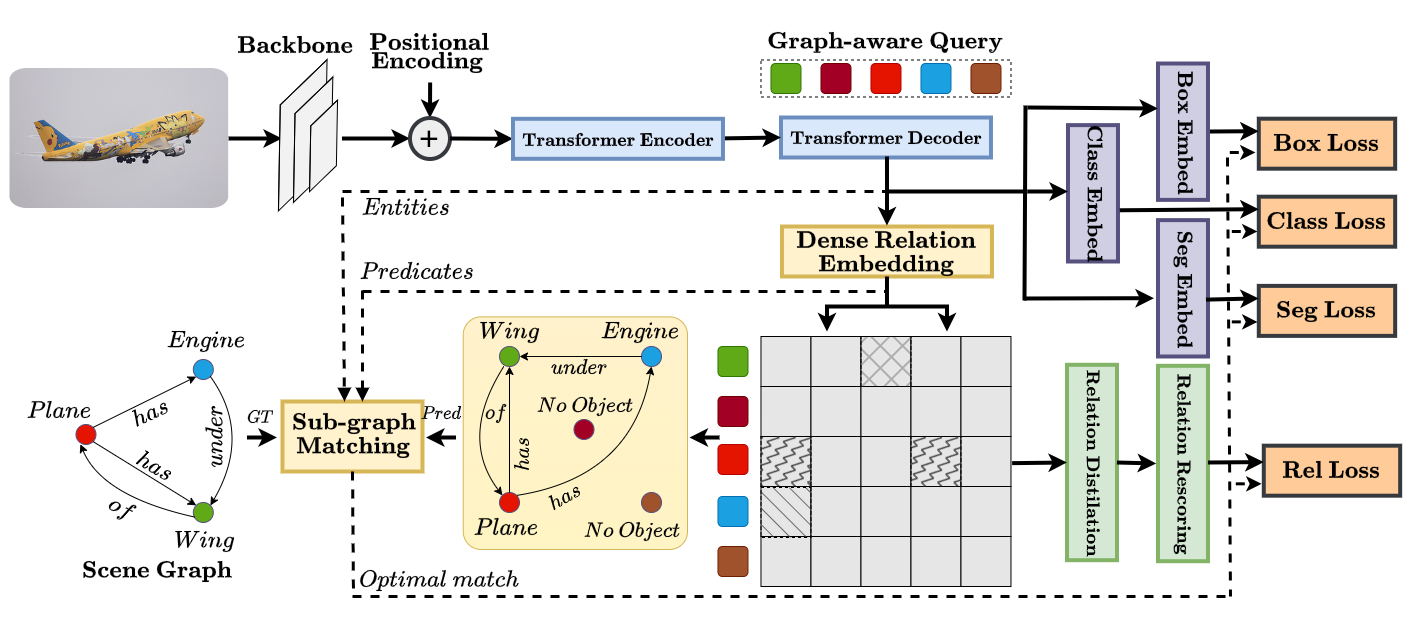}
  \end{tabular}
  \vspace{-5mm}
  \caption{An illustration of the DSGG architecture. The proposed method adopts a single-stage transformer architecture that employs graph-aware queries to predict the scene graph. The input image is first processed by the backbone network and then passed through the transformer to extract the compositional tokens. These tokens are used to learn the class confidence, bounding box, and segmentation. Additionally, a dense relation embedding module is used to learn the pairwise relation between each object in the image. A prediction graph is then generated and compared against the ground truth graph to find the optimal permutation of nodes. To rank the final relations, dense relation distillation and re-scoring modules are used. 
  }
  \label{fig:head}
  \vspace{-4mm}
\end{figure*}

\subsection{Problem Setup}\label{sec:setup}
In this section, we first introduce the problem settings for the generic scene graph generation task. Given an image $\mathcal{I} \in \mathbb{R}^{H \times W \times 3}$, our objective is to construct a comprehensive graph $\mathcal{G}$ that encapsulates the visual relationships and contextual associations in the provided image. The graph's nodes are a collection of entities $\mathcal{V}$ in the image, and the edges in the graph reflect the visual relationships $\mathcal{E}$ among these nodes. The graph is thus symbolically represented as $\mathcal{G} = (\mathcal{V}, \mathcal{E})$. Let the total number of nodes in the graph be $M$. In the scene graph generation task, the representation of each entity $\mathcal{V}_i = (c_i, b_i)$ involves its object category $c_i$, and a bounding box $b_i \in [0, 1]^4$. In contrast, the panoptic scene graph generation task introduces an additional attribute, namely, a pixel-wise semantic segmentation mask $m_i \in \mathbb{R}^{H \times W}$, to characterize each entity, as $\mathcal{V}_i = (c_i, b_i, m_i)$.
The unique object categories are denoted as $\mathcal{C} \in \{c_1, c_2, ..., c_C\}$, where $C$ represents the total number of object categories in the dataset. Moreover, edge vector $\mathcal{E}_{i,j} = (\mathcal{V}_i, r_{i,j}, \mathcal{V}_j)$ within the graph symbolize the relationships among the entities $\mathcal{V}_i$ and $\mathcal{V}_j$, often denoted as subject and object respectively. The set of relationships among all the entities is called a predicate set, denoted as $\mathcal{R} \in \{r_1, r_2, ..., r_P\}$, where $P$ is the number of unique relations. 
The list of all the edges in an image is often called a relational triplet set $\mathcal{T}$. This triplet set can be described by an adjacency matrix as $\mathcal{E} \in \mathbb{R}^{M \times M \times P}$. Note that all these edges are directed and the graph $\mathcal{G}$ can have multiple edges among the same pair of entities. The overall process involves correctly predicting this set $\mathcal{T}$. During inference, the trained models are expected to predict this ground truth triplet set using just top-$K$ predicted triplets, denoted as $\hat{\mathcal{T}}$.

\subsection{Model Architecture}
\label{sec:method}
In this section, we present our proposed DSGG network for scene graph generation, illustrated in \cref{fig:head}. We formulate the scene graph generation as a directed graph prediction task, where we denote the ground truth graph as $\mathcal{G}$ and the predicted graph as $\hat{\mathcal{G}}$. 

Our DSGG network is a single-stage transformer-based network consisting of a backbone network and an encoder-decoder network module. 
In particular, the image undergoes an initial forward pass through a backbone network, utilizing ResNet-50, Swin-B, or Swin-L. Subsequently, positional encoding is added to the image features before they undergo processing through a transformer encoder. 
We introduce a graph-aware query $\mathcal{Q}$, to extract the full graph structure from the transformer's output tokens. This query learns the object features as well as all the outgoing edge features for an object. The motivation and further details are in Sec.~\ref{sec:dense_graph}. The transformer layers are used to learn these graph-aware queries. Let $N$ be the total number of graph-aware queries in the network. Each query $\mathcal{Q}_i$ represents only the directed edges starting from a node to all other nodes (including to itself), these can be used to learn the node-specific attributes such as object class, bounding box as well as the segmentation mask directly, denoted as $\hat{\mathcal{V}_i} = (\hat{c_i}, \hat{b_i}, \hat{m_i})$. 
However, the information represented by each individual token is not enough to learn the full relationship with other objects. We generate the compositional tokens $\mathcal{S} \in \mathbb{R}^{N \times N \times P}$ on the fly which are created in a pairwise manner. The details on the compositional token embedding are in Sec.~\ref{sec:dense_graph}. The $\mathcal{S}$ are further forwarded through a series of MLP layers to learn the pairwise relation embeddings. These embeddings are further forwarded through a sigmoid layer to generate a probability map, representing the estimated dense edges of the graph $\hat{\mathcal{E}}$. The estimated $\hat{\mathcal{V}}$ and $\hat{\mathcal{E}}$ are further combined to generate the final predicted graph $\hat{\mathcal{G}}$.

The objective function is to have a perfect matching among all the nodes in the graph. However, this matching can only be attained when all the node attributes ($c_i$, $b_i$, and $m_i$) match as well as their relation triplets. 
Specifically, we aim to approximate the relation probability in \cref{eq:1} for the predicted graph $\hat{\mathcal{G}}$,
\vspace{-1mm}
\begin{equation}
\begin{aligned}
	p(\hat{\mathcal{V}_i}, \hat{\mathcal{V}_j}, \hat{\mathcal{E}_{ij}}) &= p(\hat{\mathcal{E}_{ij}} | \mathcal{Q}_i, \mathcal{Q}_j) 
\end{aligned}
\label{eq:1}
\vspace{-2mm}
\end{equation}
where $p(\hat{\mathcal{E}_{ij}} | \mathcal{Q}_i, \mathcal{Q}_j)$ denotes the probability of the predicate given the graph-aware queries $\mathcal{Q}_i$ and $\mathcal{Q}_j$. 

The number of nodes in the predicted graph $\hat{\mathcal{G}}$ and its ground truth counterpart $\mathcal{G}$ may vary. Therefore, we adopted a sub-graph matching approach to ascertain the correspondence between nodes in the predicted graph and nodes in the ground truth graph. Note that sub-graph matching is an NP-hard problem so we approximated this with an upperbound. The details on the relaxed sub-graph matching approach are in the Sec.~\ref{sec:graph_match}. The sub-graph matching provides an approximate match between the nodes in $\hat{\mathcal{G}}$ and $\mathcal{G}$. These match indices are further passed to the loss functions for estimating the gradient for the optimizer, which is detailed in Sec.~\ref{sec:method_train}. 

\subsection{Dense Scene Graph Generation}
\label{sec:dense_graph}
The main motivation behind predicting a dense graph is that it can capture a more structured representation of the objects in the images. The learned contextual information may allow the model to learn a globally consistent graph by considering relationships between different objects thus making a more informed prediction of the structured scene. This coherent representation can help mitigate the long-tail relations issues of the scene graph datasets. 

To this end, we propose to learn a dense pairwise relation representation, which can generate visual relationship triplets among any node in the graph with higher coverage in an efficient manner. A straightforward method would be to augment the number of queries in the transformer network to align with the dense relations. However, this approach introduces significant computational and memory complexity, particularly for transformer-based networks known for their resource-intensive nature. To tackle this challenge, we introduce a set of graph-aware queries that learn a compact representation of the object's attributes and its relations with all other objects in the image.
This differs from existing approaches that predict only a relatively small set of triplets $\hat{\mathcal{T}}$ for each image, 
which fall short of capturing the full spectrum of relationships among objects within the graph.

\paragraph{Graph-aware Queries:}
We first introduce our graph-aware queries, which can be used to generate compositional tokens efficiently for dense graph prediction. Specifically, the representation of each predicted object is combined with the representations of all other objects, aiming to approximate the dense relational embedding between every pair of nodes in the graph as shown below,
\vspace{-2mm}
\begin{equation}
	p(\hat{\mathcal{E}_{ij}} | \mathcal{Q}_i, \mathcal{Q}_j) = MLP(\mathcal{Q}_i \bigoplus \mathcal{Q}_j) 
 \label{eq:2}
 \vspace{-2mm}
\end{equation}
Note that $\bigoplus$ is a concatenation operator, thus learning a directed graph. The MLP layers are trained to learn multi-label relations $\hat{\mathcal{E}} \in \mathbb{R}^{N \times N \times P}$. A \textit{Sigmoid} activation function is used to learn the probability for each predicate label. This is in contrast to triplet-based approaches, which employ a \textit{Softmax} activation and can only learn a single relation ($\hat{\mathcal{E}} \in \mathcal{R}^{M \times P}$) for each learned query. we learn the conditional probability in Equ.~\ref{eq:2} with a two-step approach. 

\paragraph{Relation Distillation:}
\label{sec:method_rel_dist}
First, we train a predicate filter to dynamically reject pairwise relations. Note that, during training we do not use a predefined set of possible triplets based on training dataset statistics. Instead, we learn it from the data to capture the missing triplets that may occur in the future. The structured graph learning approach has the potential to leverage related triplets in the images. Specifically, this predicate filter is trained using graph-aware queries. It takes the $\mathcal{Q}_i$ as input and computes a cross-attention with all other $\mathcal{Q}_j$ to approximate a binary adjacency matrix. However, we also replaced the \textit{Softmax} operator in the attention with a \textit{Sigmoid} function to attend to multiple queries.
This relation filter $\mathcal{F} \in \mathbb{R}^{N\times N}$ is learned as follows. 
\vspace{-2mm}
\begin{equation}
	\mathcal{F} = sigmoid(\frac{\mathcal{Q} \cdot \mathcal{Q}^{T}}{\sqrt{d_{q}}})
\label{eq:3}
\vspace{-2mm}
\end{equation}
This filter learns if there exists at least a single relation for each entity pair rather than finding the exact relation.
Moreover, along with the relation filter, we learn an MLP based on pairwise features to dynamically distill relations.  
\vspace{-2mm}
\begin{equation}
	p(\hat{\mathcal{E}_{ij}} | \mathcal{S}_{i,j}) = \mathcal{F} \cdot p(\hat{\mathcal{E}_{ij}} | \mathcal{Q}_i, \mathcal{Q}_j) 
 \label{eq:4}
 \vspace{-2mm}
\end{equation}

\paragraph{Relation Re-scoring}
\label{sec:method_rel_rescore}
Secondly, to reward entity queries with a higher object confidence, we apply a relation re-scoring strategy. It's important to note that our proposed approach is not constrained by the requirement that the number of predicates matches the number of relations in the transformer queries. Also, it is possible that a high-scoring object can have no relations. ~\cite{zhang2022upt} adapted a lambda scaling approach to suppress the overconfident objects. Inspired by ~\cite{zhang2022upt}, we adopt a similar re-scoring mechanism and extend it to rank all the pairwise relations. However, since all the confidences are learned in an end-to-end manner, the DSGG doesn't require lambda scaling explicitly. Specifically, the final pairwise relation probability is calculated as follows,
\vspace{-2mm}
\begin{equation}
\begin{aligned}
	p(\hat{\mathcal{E}_{ij}} | \mathcal{S}_{i,j}) &= p(\hat{\mathcal{V}_{i}} | \mathcal{Q}_i) \cdot p(\hat{\mathcal{V}_{j}} | \mathcal{Q}_j) \cdot p(\hat{\mathcal{E}_{ij}} | \mathcal{Q}_i, \mathcal{Q}_j) \\
\end{aligned}
\label{eq:5}
\vspace{-2mm}
\end{equation}
where $p(\hat{\mathcal{V}_{i}})$ is learned using the corresponding embeddings, encompassing ($c_i$, $b_i$, and $m_i$).  
We use the same re-scoring module as in Equ.~\ref{eq:5} for both the scene-graph generation and the panoptic scene-graph generation tasks.

\paragraph{Logits Adjustment:} We employed~\cite{sadeep_logits} as a post-processing step to mitigate bias in the features resulting from relational noise. Specifically, we set the logits adjustment weight to 0.15. Moreover, we implemented an NMS-based strategy to eliminate duplicate triplets, relying on bounding boxes for the SGG task and segmentation masks for the PSG task. Finally, only the top 100 triplets are used to evaluate our method.

\subsection{Sub-graph Matching}
\label{sec:graph_match}
Given the generated graph, we now introduce our graph matching procedure for identifying a correspondence between the prediction $\hat{\mathcal{G}}$ and the ground truth graph $\mathcal{G}$, with $N$ and $M$ nodes respectively.  
Without loss of generality, we suppose that both graphs have the same order of $N$. Otherwise, they could be expanded with new dummy nodes that have no relation to any other node in the graph. Specifically, we add $N-M$ dummy nodes to make both graphs have the same number of nodes. 
Similar to the nodes and edges of $G$ defined in Sec.~\ref{sec:setup}, we denote the nodes and edges in the predicted graph $\hat{\mathcal{G}}$ as $\{\hat{\mathcal{V}}_i\}_{i\in \hat{\mathcal{V}}_{\hat{\mathcal{G}}}}$ and $\{\hat{r}_{i,j}\}_{(i,j)\in \mathcal{E}_{\hat{\mathcal{G}}}}$, respectively.  

A correspondence $\sigma_{i,a}$ between the $i^{th}$ node of $\mathcal{G}$ and the $a^{th}$ node of $\hat{\mathcal{G}}$ is a bijective function that assigns one node of $\mathcal{G}$ to only one node of $\hat{\mathcal{G}}$. We represent the optimal mapping from graph $\mathcal{G}$ to graph $\hat{\mathcal{G}}$ as $\hat{\sigma}$. Then, a generic formulation of the graph matching problem consists of finding the optimal correspondence $\hat{\sigma}$ given by the solution of a quadratic assignment (QA) problem~\cite{kurt2003}: 
\vspace{-2mm}
\begin{equation}
\label{eq:matching}
    \hat{\sigma} = \underset{\forall \sigma: \mathcal{G} \to \hat{\mathcal{G}}}{\arg\min} \biggl\{ \sum_{\forall i} C_{e}(\mathcal{V}_i, \hat{\mathcal{V}}_{\sigma(i)}) +  \sum_{\forall {(i,j)}}  C_{r}(r_{i,j}, \hat{r}_{\sigma(i,j)}) \biggl\},
\end{equation}
where $C_{e}(\mathcal{V}_i, \hat{\mathcal{V}}_{\sigma(i)})$ is a pair-wise \emph{entity matching cost} between ground truth node $i$ and a prediction with index $\sigma(i)$. Similarly, $C_{r}(r_{i,j}, \hat{r}_{\sigma(i,j)})$ is a pair-wise \emph{relation matching cost} between ground truth relation $r_{i,j}$ and a prediction with index $\sigma(i, j)$. 
In our problem, the matching cost takes into account the class prediction, the similarity of predicted and ground truth boxes, and the similarity of predicted and ground truth relations.

Specifically, we define
$C_{e}(\mathcal{V}_i, \hat{\mathcal{V}}_{\sigma(i)})$ as $\mathbbm{1}_{c_i\neq\emptyset}(1-\hat{p}_{\sigma(i)}(c_i)) + \mathbbm{1}_{c_i\neq \emptyset} \text{IoU}({b_{i},\hat{b}_{\sigma(i)}})$, where $\hat{p}_{\sigma(i)}$ is the predicted entity class probability. Similarly, we define $C_{r}(r_{i,j}, \hat{r}_{\sigma(i,j)})$ as $\mathbbm{1}_{c_i,c_j\neq\emptyset}(1-\hat{p}_{\sigma(i,j)})^{\top}r_{i,j} + \mathbbm{1}_{c_i,c_j = \emptyset}(\hat{p}_{\sigma(i,j)})^{\top}(1-r_{i,j})$, where $\hat{p}_{\sigma(i,j)}$ is predicted relation class probability vector.

As the generic QA problem is NP-hard~\cite{kurt2003}, we introduce an approximation scheme to reduce the computation cost. In particular, we upper-bound the quadratic term as below: 
\begin{equation}
\begin{aligned}
\label{eq:matching2}
     \sum_{\forall (i,j)} C_{r}\leq \biggl\{ \sum_{\forall i} \sum_{\forall j} \mathbbm{1}_{c_i,c_j\neq\emptyset}(1-\hat{p}_{\sigma(i),j})^{\top}r_{i,j} \\ + \sum_{\forall i} \sum_{\forall j} \mathbbm{1}_{c_i,c_j = \emptyset}(\hat{p}_{\sigma(i),j})^{\top}(1 - r_{i,j}) \biggl\},
\end{aligned}
\end{equation}
This reduces the QA problem to a linear form and its optimal assignment can be computed efficiently with the Hungarian algorithm, following prior work (\eg ~\cite{kuhn1955hungarian}). We note that this approximation also allows us to train this dense relation layer jointly with an end-to-end transformer model, as detailed in the following subsection.

\subsection{End-to-end Graph Learning}
\label{sec:method_train}
For the scene graph generation task, we employ DETR with query denoising~\cite{li2022dn} and train $N$ graph-aware queries. Conversely, in the panoptic scene graph generation task, we adhere to the settings outlined in~\cite{yang2022panoptic,cheng2021mask2former} and also train $N$ graph-aware queries. We adopt the sub-graph matching, as discussed in~\cref{sec:graph_match}, to find the best node assignment in the predicted graph. During training, overall objective is based on a multi-task loss, which contains the bounding box L1 loss $\mathcal{L}_{box}$, the GIoU loss $\mathcal{L}_{giou}$, the entity classification loss $\mathcal{L}_{ent}$, and the pairwise relation loss $\mathcal{L}_{rel}$. 
This multi-task loss $\mathcal{L}$ is formulated as:
\vspace{-1mm}
\begin{equation}
    \mathcal{L} = \lambda_{box}\mathcal{L}_{box} + \lambda_{giou}\mathcal{L}_{giou} + \lambda_{ent}\mathcal{L}_{ent} +  \lambda_{rel}\mathcal{L}_{rel}, 
    \vspace{-1mm}
\end{equation}
where $\mathcal{L}_{rel}$ refers to a focal loss that is applied to all pairwise edges in the extended graph. The hyper-parameters used to weight each loss are $\lambda_{box}$, $\lambda_{giou}$, $\lambda_{ent}$, and $\lambda_{rel}$. 
We also trained segmentation focal and dice losses, $\mathcal{L}_{dice}$ and $\mathcal{L}_{focal}$, with $\lambda_{dice}=\lambda_{focal}=1$ for panoptic scene graph generation. In addition to this, we improve the relation accuracy by leveraging logit adjustment at the time of inference. Note that the DSGG model is trained in an end-to-end manner for both scene graph and panoptic scene graph generation tasks.

\section{Experiments}
\label{sec:exp}
In this section, we demonstrate the effectiveness of our method on scene graph generation (SGG) and panoptic scene graph generation (PSG) tasks. 

\begin{table*}[t]
\begin{adjustbox}{width=2.1\columnwidth,center}
\begin{tabular}{rccccccccc}
\toprule
~ & \multicolumn{3}{c}{\textbf{Predicate Classification (PredCIs)}} & \multicolumn{3}{c}{\textbf{Scene Graph Classification (SGCIs)}} & \multicolumn{3}{c}{\textbf{Scene Graph Detection (SGDet)}} \\
\textbf{Method}      & \textbf{R@50/100} & \textbf{mR@50/100} & \textbf{M@50/100} & \textbf{R@50/100} & \textbf{mR@50/100} & \textbf{M@50/100} & \textbf{R@50/100} & \textbf{mR@50/100} & \textbf{M@50/100} \\ 
\midrule
Motifs~\cite{zellers2018scenegraphs}       & 65.3 / 67.2 & 14.9 / 16.3 & 40.1 / 41.8    & 38.9 / 39.8 & 8.3 / 8.8 & 23.6 / 24.3       & 32.1 / 36.8 & 6.6 / 7.9 & 19.4 / 22.4      \\ 
VCTree~\cite{tang2020unbiased}   & 65.5 / 67.4 & 16.7 / 17.9 & 41.1 / 42.7    & 40.3 / 41.6 & 7.9 / 8.3 & 24.1 / 25.0       & 31.9 / 36.0 & 6.4 / 7.3 & 19.2 / 21.7      \\ 
BGNN~\cite{Li_2021_CVPR}   & 59.2 / 61.3 & 30.4 / 32.9 & 44.8 / 47.1    & 37.4 / 38.5 & 14.3 / 16.5 & 25.9 / 27.5     & 31.0 / 35.8 & 10.7 / 12.6 & 20.9 / 24.2    \\ 
SGTR~\cite{Li_2022_CVPR}          & - & - & -                                  & - & - & -                                   & 20.6 / 25.0 & 15.8 / 20.1    & 18.2 / 22.6 \\ 
SS-RCNN~\cite{Teng_2022_CVPR} & - & - & -                                  & - & - & -                     & 23.7 / 27.3 & \underline{18.6} / \underline{22.5} & 21.2 / 24.9 \\
SHA-GCL~\cite{dong2022stacked}       & 35.1 / 37.2 & \textbf{41.6} / \underline{44.1} & 38.4 / 40.7    & 22.8 / 23.9 & \underline{23.0} / 24.3 & 22.9 / 24.1     & 14.9 / 18.2 & 17.9 / 20.9    & 16.4 / 19.6    \\ 
NICE~\cite{Li0HZZ022} & 55.1 / 57.2 & 29.9 / 32.3 & 42.5 / 44.8    & 33.1 / 34.0 & 16.6 / 17.9 & 24.9 / 26.0     & 27.8 / 31.8 & 12.2 / 14.4    & 20.0 / 23.1    \\  
HL-Net~\cite{lin2022hl}  & \underline{67.0} / 68.9 & - / 22.8 & - / 45.9          & \textbf{42.6} / \textbf{43.5} & - / 13.5 & 28.5               & \textbf{33.7} / \underline{38.1} & - / 9.2 & - / 23.7 \\
RU-Net~\cite{lin2022ru}  & \textbf{67.7} / \underline{69.6} & - / 24.2 & - / 46.9          & \underline{42.4} / \underline{43.3} & - / 14.6 & - / 29.0           & 32.9 / 37.5 & - / 10.8    & - / 24.2    \\ 
Relationformer~\cite{Relationformer} & - & - & - & - & - & - & 28.4 / 31.3 & 9.3 / 10.7 & 18.9 / 21.0 \\
RepSGG~\cite{liu2023repsgg} & 27.8 / 28.8 & \underline{39.7} / 43.7 & 33.8 / 36.3 & 17.9 / 20.3 & 22.3 / 27.7 & 20.1 / 24.0 & 12.1 / 14.6 & 15.3 / 18.9 & 13.7 / 16.8 \\
HetSGG~\cite{HetSGG}   & 57.8 / 59.1 & 31.6 / 33.5 & 44.7 / 46.3    & 37.6 / 38.7 & 17.2 / 18.7 & 27.4 / 28.7     & 30.0 / 34.6 & 12.2 / 14.4    & 21.1 / 24.5    \\ 
PE-Net~\cite{zheng2023prototype} & 59.0 / 61.4 & 38.8 / 40.7 & \textbf{48.9} / 51.1 & 36.1 / 37.3 & 22.2 / 23.5 & 29.2 / 30.4   & 26.5 / 30.9 & 16.7 / 18.8    & 21.6 / 24.9    \\ 
\midrule
\textbf{DSGG (ours)} $^\dagger$ & 65.3 / \textbf{75.0} & 31.2 / 41.6 & \underline{48.3} / \textbf{58.3}        & 38.8 / 41.4 & 19.9 / \underline{25.0} & \textbf{29.4} / \underline{33.2}      & \underline{32.9} / \bf{38.5} & 13.0 / 17.3 & \underline{23.0} / \underline{28.0}    \\ 
\textbf{DSGG (ours)}  & 53.9 / 65.1 & 39.4 / \textbf{49.9} & 46.7 / \underline{57.5}        & 33.1 \ 38.0 & \textbf{23.7} / \textbf{29.7} & \underline{28.4} / \textbf{33.9}           & 26.5 / 32.9 & \bf{20.2} / \bf{25.5} & \bf{23.4} / \bf{29.2}  \\ 
\bottomrule
\end{tabular}
\end{adjustbox}
\caption{\textbf{Evaluation on the Visual Genome dataset}~\cite{tang2020unbiased}. The {\bf best} and \underline{second} best methods under each setting are marked according to formats. $^\dagger$ shows DSGG results without logit adjustment. Comparisons to additional methods are included in the supplementary material.}
\label{tab:vg_overall}
\vspace{-4mm}
\end{table*}

\subsection{Datasets}
To evaluate the effectiveness of our approach, we conduct experiments on the widely recognized and challenging datasets, namely Visual Genome~\cite{tang2020unbiased} and PSG~\cite{yang2022panoptic} dataset. 
Both these scene graph datasets contain a list of <subject, predicate, object> triplets that are often noisy and duplicated. Additionally, there are several concurrent relationships for objects, or sets of objects, pointing to a relational semantic overlap problem.
The following are each dataset's statistics.

\paragraph{Visual Genome (VG)}~\cite{tang2020unbiased} dataset has {\tt 108,077} images and their associated scene graph annotations featuring {\tt 50} predicate relationships and {\tt 150} object categories.  We followed~\cite{XuZCF17} for the training, validation, and test splits. 

\paragraph{Panoptic Scene Graph (PSG)}~\cite{yang2022panoptic} dataset has {\tt 48,749} images, {\tt 80} thing classes, {\tt 53} stuff categories and {\tt 56} predicate relationships. The dataset contains {\tt 2,177} test images, with {\tt 28} rare predicates and {\tt 28} non-rare relation categories. There are multiple relationships among the same objects in {\tt 927} images in the test split. We followed~\cite{yang2022panoptic} for the training and the test splits. 

\subsection{Evaluation Metrics}
Following~\cite{tang2020unbiased, zheng2023prototype}, we report recall ({\bf R}), mean recall ({\bf mR}) and overall mean {\bf M@K} accuracy (\%) on the test set of Visual Genome and Panoptic scene graph generation datasets. We also report predicate classification (PredCIs), scene graph classification (SGCIs), and scene graph detection (SGDet) metrics for the Visual Genome dataset.

\subsection{Implementation Details}
The DSGG models are trained with 100 graph-aware queries only. Specifically, we use the ResNet-50~\cite{li2022dn}, Swin-B~\cite{liu2021swin} and Swin-L~\cite{liu2021swin} as the backbone networks. For the SGG task, we trained the DSGG model for 60 epochs only, as in~\cite{tang2020sggcode}. Note that our models are trained from scratch without a pre-trained object detector on the VG dataset. For the PSG task, we followed~\cite{yang2022panoptic} and, in order to provide a fair comparison to the baselines, we only fine-tuned our model initialized with Mask2Former model~\cite{cheng2021mask2former} (pre-trained on the COCO~\cite{lin2014microsoft} dataset) for 12-epochs. In both settings, we jointly train the shared transformer's encoder and decoder for learning compositional query tokens. These 256-dimensional tokens are then forwarded to class-embedding and box-embedding for learning the labels and bounding boxes for the SGG task. Additionally, we use the segmentation embedding to learn the object's pixel-wise semantic segmentation for the PSG task only. The number of encoder and decoder layers is kept as default and we adopted the same data augmentation settings as in the baselines~\cite{tang2020sggcode, yang2022panoptic}. The models are trained end-to-end with the sub-graph matching as the default cost function for both SGG and PSG tasks. AdamW~\cite{loshchilov2018decoupled} is used as an optimizer with a weight decay of $10^{-4}$. We set the initial learning rate of the backbone, transformer, and scene-graph generation to $10^{-5}$, $10^{-4}$, and $10^{-4}$ respectively. Four A100 GPUs are used for both training and evaluating the models; however, for the PSG and SGG tasks, we used batches of 1 and 4 images, respectively.

\subsection{Experimental Evaluation} 

\paragraph{Scene Graph Generation:} 
Table~\ref{tab:vg_overall} shows scene graph detection results on the test split of the Visual Genome dataset~\cite{tang2020unbiased}. Our method achieves state-of-the-art recall without the need for any complex post-processing of scene graphs. Note that in this setting, the relation prediction is restricted to unbiased scene graph generation. However, following~\cite{Teng_2022_CVPR}, we also applied logit adjustment (LA) as a post-processing step to fix the long-tailed issue with relations categories in the dataset. Note that this approach outperforms all the baselines, by a considerable margin in the case of mean recall and the mean@K metric. Specifically, we compare our method with Motifs~\cite{zellers2018scenegraphs}, Unbiased~\cite{tang2020unbiased}, BGNN~\cite{Li_2021_CVPR}, SGTR~\cite{Li_2022_CVPR}, Structured Sparse RCNN~\cite{Teng_2022_CVPR}, SHA-GCL~\cite{dong2022stacked}, NICE~\cite{Li0HZZ022}, HL-Net~\cite{lin2022hl}, RU-Net~\cite{lin2022ru}, Relationformer~\cite{Relationformer}, RepSGG~\cite{liu2023repsgg}, HetSGG~\cite{HetSGG}, and PE-Net~\cite{zheng2023prototype}. Note also that our approach with the LA post-processing is competitive in terms of recall. 

\begin{table}[h!]
\begin{adjustbox}{width=\columnwidth,center}
    \begin{tabular}{rccccccc}
    \toprule
        ~ & ~ & \multicolumn{5}{c}{\textbf{Panoptic Scene Graph Detection}} \\
        \cmidrule{3-8}
        \textbf{Method} & \textbf{Backbone} & \textbf{R@20} & \textbf{mR@20} & \textbf{R@50} & \textbf{mR@50} & \textbf{R@100} & \textbf{mR@100} \\
        \toprule
        IMP \cite{XuZCF17} & R50 & 16.5 & 6.5 & 18.2 & 7.1 & 18.6 & 7.2 \\
        MOTIF \cite{zellers2018scenegraphs} & R50 & 20.0 & 9.1 & 21.7 & 9.6 & 22.0 & 9.7 \\
        VCTree \cite{tang2020unbiased} & R50 & 20.6 & 9.7 & 22.1 & 10.2 & 22.5 & 10.2 \\
        GPSNet \cite{lin2020gps} & R50 & 17.8 & 7.0 & 19.6 & 7.5 & 20.1 & 7.7 \\
        \midrule
        PSGTR \cite{yang2022panoptic} & R50 & 28.4 & 16.6 & 34.4 & 20.8 & 36.3 & 22.1 \\
        PSGFormer \cite{yang2022panoptic} & R50 & 18.0 & 14.8 & 19.6 & 17.0 & 20.1 & 17.6 \\
        HiLo \cite{zhou2023hilo} $^\dagger$ & R50 & \textbf{34.1} & \underline{23.7} & \underline{40.7} & \underline{30.3} & \underline{43.0} & \underline{33.1} \\
        \textbf{DSGG (ours)} & R50 & \underline{32.7} & \textbf{30.8} & \textbf{42.8} & \textbf{38.8} & \textbf{50.0} & \textbf{43.4} \\
        \midrule
        HiLo \cite{zhou2023hilo} $^\dagger$ & Swin-B & \textbf{38.5} & \underline{28.3} & \underline{46.2} & \underline{35.3} & \underline{49.6} & \underline{39.1} \\
        \textbf{DSGG (ours)} & Swin-B & \underline{35.5} & \textbf{32.9} & \textbf{46.5} & \textbf{41.3} & \textbf{54.2} & \textbf{46.3} \\
        \midrule
        HiLo \cite{zhou2023hilo} $^\dagger$ & Swin-L & \textbf{40.6} & \underline{29.7} & \textbf{48.7} & \underline{37.6} & \underline{51.4} & \underline{40.9} \\
        \textbf{DSGG (ours)} & Swin-L & \underline{36.2} & \textbf{34.0} & \underline{47.0} & \textbf{41.7} & \textbf{54.3} & \textbf{47.8} \\
        \midrule
        \textbf{DSGG (ours)} $^\dagger$ & R50 & 32.2 & 30.9 & 42.5 & 40.1 & 49.7 & 44.1 \\
        \textbf{DSGG (ours)} $^\dagger$ & Swin-B & 35.8 & 33.9 & 46.3 & 43.2 & 54.5 & 48.7 \\
        \textbf{DSGG (ours)} $^\dagger$ & Swin-L & 36.0 & 34.1 & 47.0 & 42.1 & 54.7 & 48.0 \\
        \bottomrule
    \end{tabular}
\end{adjustbox}
\caption{\textbf{Evaluation on the PSG dataset}~\cite{yang2022panoptic}. The {\bf best} and \underline{second} best methods under each setting are marked according to formats. $^\dagger$ represents the models trained using additional relation labels obtained through a baseline-trained model. } 
\label{table:psg_main_results} 
\vspace{-4mm}
\end{table}

\paragraph{Panoptic Scene Graph Generation:}
We followed~\cite{yang2022panoptic, zhou2023hilo} for the evaluation of recall and mean recall (\%) using the panoptic segmentation as the default criteria. Table~\ref{table:psg_main_results} shows performance comparision of DSGG with several baselines~\cite{XuZCF17, zellers2018scenegraphs, tang2020unbiased, lin2020gps, yang2022panoptic, zhou2023hilo}. Our method achieves a consistent improvement over all the metrics. An analysis of the top-20 relations, the method proposed by ~\cite{zhou2023hilo} demonstrates slightly improved performance, yet the recall metric is heavily influenced by high-frequency relations. Note that, their approach yields considerably lower results for the mean-recall metric, which assigns equal weight to all relation classes, and is a better overall metric for scene-graph comparison. The DSGG attains superior results for both metrics across the top 50 and 100 relations on PSG dataset.

\subsection{Ablation Studies}
In this section, we provide a comprehensive analysis of our model, emphasizing various aspects. Initially, we conducted ablation experiments to evaluate the contributions of individual components in the DSGG model. Furthermore, we perform an ablation study that examines performance using top-scene graph predictions and explores the zero-shot capabilities inherent in our model.

\paragraph{Effectiveness of different components of the model:} In this ablation study, we explore how the different components of the model influence the final performance on the VG dataset. In particular, our attention is directed towards understanding the effectiveness of the relation distillation and re-scoring mechanism, and the role of logit adjustment components. Table~\ref{tab:vg_comp} demonstrates the outcomes for various combinations of these components. 
Note that our graph-aware queries learn relation triplets more effectively and thus contribute to a significant improvement in the overall performance. The recall is improved by the relation rescoring and distillation modules. However, logit adjustment yields better mean recall and mean performance overall.

\begin{table}[!b]
\vspace{-2mm}
\begin{adjustbox}{width=\linewidth,center}
\begin{tabular}{cccccc}
\toprule
\textbf{Relation} & \textbf{Relation} & \textbf{Logits} & \textbf{Recall (Main)}    & \textbf{Mean Recall} & \textbf{Mean @ K} \\
\textbf{Rescoring} & \textbf{Distillation} & \textbf{Adjustment} & \textbf{R@50/100} & \textbf{mR@50/100} & \textbf{M@50/100} \\ 
\midrule
 & & & 6.9/11.2         & 11.9/15.6           & 9.4/13.4        \\  
$\checkmark$ &  & & 25.9/32.6 & 12.5/15.7 & 19.2/24.2 \\
$\checkmark$ & $\checkmark$ & & \bf{32.9/38.5} & 13.0/17.3 & 23.0/28.0        \\
$\checkmark$ & $\checkmark$ & $\checkmark$  & 26.5/32.9 & \bf{20.2/25.5} & \bf{23.4/29.2}\\
\bottomrule
\end{tabular}
\end{adjustbox}
\vspace{-3mm}
\caption{Impact assessment of various components of the model on the scene graph detection task using the Visual Genome test set. }
\label{tab:vg_comp}
\end{table}

\paragraph{Effectiveness of the model on top predictions:} 
Table~\ref{tab:vg_20_50_zeroshot} shows the influence of the number of graph-aware queries in the transformer on the Visual Genome dataset. During the testing phase, several scene-graph generation approaches struggle with localizing objects and predicates with a smaller number of queries. Note that, our model trained with just 20 queries achieves the best results on both metrics. 

\paragraph{Effectiveness of the model on zero-shot learning: }
We also study the generalization capability of DSGG to unseen relationships. We, therefore, evaluated the zero-shot performance of our method on the Visual Genome test set. Table~\ref{tab:vg_20_50_zeroshot} shows a performance comparison with several baseline approaches. Note that, our model achieves consistently state-of-the-art results on zero-shot recall for all metrics. 

\begin{table}[!t]
\centering \small
\scalebox{0.59}{
\begin{minipage}{.45\textwidth}
\begin{tabular}{rcccc}
\toprule
\textbf{Visual Genome}          & \textbf{Mean Recall}  & \textbf{Recall}    & \textbf{Mean @ K} \\
\textbf{SGDet}                  & \textbf{mR@20}        & \textbf{R@20}             & \textbf{M@20} \\ 
\midrule
\textbf{Unbiased}~\cite{tang2020unbiased}               & 6.9           & 19.0         & 13.0        \\ 
\textbf{SS-RCNN}~\cite{Teng_2022_CVPR} & 13.7          & 18.2      & 15.8     \\
\textbf{SHA-GCL}~\cite{dong2022stacked}                & \underline{14.2}          & -         & -        \\ 
\textbf{HL-Net}~\cite{lin2022hl}                 & -             & \bf{26.0} & -        \\   
\textbf{RU-Net}~\cite{lin2022ru}                 & -             & \underline{25.7}      & -        \\
\textbf{PE-Net}~\cite{zheng2023prototype}                 & 9.2           & 23.4      & \underline{16.3}     \\ 
\midrule
\textbf{DSGG (ours)} $^\dagger$      & 8.3           & 23.4      & 15.9        \\ 
\textbf{DSGG (ours)}      & \bf{14.2}    & 18.7      & \bf{16.4}        \\
\bottomrule
\end{tabular}
\end{minipage} 
\begin{minipage}{.45\textwidth}
\begin{tabular}{rccc}
\toprule
\textbf{Visual Genome}          &  \multicolumn{2}{c}{\textbf{Zero Shot Recall}} \\
\textbf{SGDet}                  & \textbf{zR@50}             & \textbf{zR@100} \\ 
\midrule
\textbf{Motifs}~\cite{zellers2018scenegraphs}                 & 0.1       & 0.1      \\ 
\textbf{Motifs + TDE}~\cite{tang2020unbiased}                 & 2.3       & 2.9      \\   
\textbf{VCTree}~\cite{tang2020unbiased}                       & 0.3       & 0.7      \\ 
\textbf{VCTree + TDE}~\cite{tang2020unbiased}                 & 2.6       & 3.2        \\
\textbf{SS-RCNN}~\cite{Teng_2022_CVPR}                   & \underline{3.1}       & \underline{4.5}      \\
\textbf{PE-Net}~\cite{zheng2023prototype}                     & 2.3       & 3.6      \\ 
\midrule
\textbf{DSGG (ours)} $^\dagger$ & 2.5       & 3.9      \\ 
\textbf{DSGG (ours)}            & \bf{3.5}  & \bf{5.2} \\
\bottomrule
\end{tabular}
\end{minipage}
}
\caption{\textbf{Left:} Evaluation of top-20 relation triplets on the Visual Genome test set. \textbf{Right:} Evaluation of Zero-shot Recall performance on the Visual Genome test set. $^\dagger$ shows DSGG results without logit adjustment.}
\label{tab:vg_20_50_zeroshot}
\vspace{-5mm}
\end{table}

\vspace{-2mm}
\begin{table}[b!]
\begin{adjustbox}{width=\columnwidth,center}
    \begin{tabular}{rccccccc}
    \toprule
        ~ & ~ & ~ & \multicolumn{5}{c}{\textbf{Relational Semantic Overlap}} \\
        \cmidrule{3-8}
        \textbf{Method} & \textbf{Backbone} & \textbf{R@20} & \textbf{mR@20} & \textbf{R@50} & \textbf{mR@50} & \textbf{R@100} & \textbf{mR@100} \\
        \toprule
        HiLo \cite{zhou2023hilo} & R50 & 43.6 & 30.8 & 49.7 & 36.2 & 51.1 & 38.8 \\
        \textbf{DSGG (ours)} & R50 & 48.6 & 37.6 & 58.2 & 48.6 & 63.6 & 50.2 \\
        $\Delta$ & & \textcolor{teal}{+5.0} & \textcolor{teal}{+6.8} & \textcolor{teal}{+8.5} & \textcolor{teal}{+12.4} & \textcolor{teal}{+12.5} & \textcolor{teal}{+11.4} \\
        \midrule
        HiLo \cite{zhou2023hilo} & Swin-B & 51.3 & 36.4 & 57.9 & 42.2 & 59.9 & 45.0 \\
        \textbf{DSGG (ours)} & Swin-B & 52.7 & 41.3 & 60.9 & 49.6 & 66.7 & 54.7 \\
        $\Delta$ & & \textcolor{teal}{+1.4} & \textcolor{teal}{+4.9} & \textcolor{teal}{+3.0} & \textcolor{teal}{+7.4} & \textcolor{teal}{+6.8} & \textcolor{teal}{+9.7} \\
        \midrule
        HiLo \cite{zhou2023hilo} & Swin-L & 53.1 & 36.3 & 60.7 & 46.7 & 62.6 & 49.0 \\
        \textbf{DSGG (ours)} & Swin-L & 53.4 & 42.6 & 62.1 & 50.3 & 68.0 & 55.2 \\
        $\Delta$ & & \textcolor{teal}{+0.3} & \textcolor{teal}{+6.3} & \textcolor{teal}{+1.4} & \textcolor{teal}{+3.6} & \textcolor{teal}{+5.4} & \textcolor{teal}{+6.2} \\
        \bottomrule
    \end{tabular}
\end{adjustbox}
\vspace{-2mm}
\caption{DSGG and~\cite{zhou2023hilo} performance comparison on the PSG dataset's relational semantic overlap subset. This subset consists of images that show various relationships between the same objects. Our approach consistently outperforms HiLo~\cite{zhou2023hilo}.} 
\label{table:psg_soi_results} 
\end{table}

\subsection{Analysis}
This section outlines the analysis of the proposed DSGG model concerning challenges related to long-tail distribution and relational semantic overlap. All the experiments are carried out on the PSG dataset.

\paragraph{Relational Semantic Overlap:}
The issue of relational semantic overlap arises when there are several relationships between the same pair of entities. For instance, a typical example is an image where a person is simultaneously holding a horse and looking at it. In this specific section, we focus on assessing the effectiveness of the proposed DSGG method and the baseline approaches when dealing with images that have entities exhibiting relational semantic overlap. The PSG test dataset consists of 927 images that depict multiple relationships between the same entities. Table~\ref{table:psg_soi_results} shows a performance comparison of the proposed DSGG method and the current state-of-the-art HiLo~\cite{zhou2023hilo} approach, it is evident that the proposed DSGG method excels at effectively addressing the challenge of relational semantic overlap.

\paragraph{Low-frequency Relations:}
A relation category is deemed rare in the PSG dataset if it encompasses fewer than 500 instances. Table~\ref{table:psg_longtail_results} shows a performance comparison of the proposed DSGG method and the current state-of-the-art HiLo~\cite{zhou2023hilo} approach. The proposed DSGG method is robust to the rare relation categories and consistently excels at effectively predicting low-frequency relations in the presence of high-frequency predicate categories.

\begin{table}[t!]
\begin{adjustbox}{width=1\columnwidth,center}
    \begin{tabular}{rcccc}
    \toprule
        ~ & ~ & \multicolumn{3}{c}{\textbf{Low-frequency Relations (28)}} \\
        \cmidrule{3-5}
        \textbf{Method} & \textbf{Backbone} & \textbf{mR@20} & \textbf{mR@50} & \textbf{mR@100} \\
        \toprule
        HiLo \cite{zhou2023hilo} & R50 & 10.4 & 17.0 & 20.3 \\
        \textbf{DSGG (ours)} & R50 & 20.9 \textcolor{teal}{(+10.5)} & 31.0 \textcolor{teal}{(+14.0)} & 33.7 \textcolor{teal}{(+13.4)} \\
        \midrule
        HiLo \cite{zhou2023hilo} & Swin-B & 13.1 & 20.3 & 24.6 \\
        \textbf{DSGG (ours)} & Swin-B & 23.0 \textcolor{teal}{(+9.9)} & 33.9 \textcolor{teal}{(+13.6)} & 38.1 \textcolor{teal}{(+13.5)} \\
        \midrule
        HiLo \cite{zhou2023hilo} & Swin-L & 14.2 & 22.6 & 26.3 \\
        \textbf{DSGG (ours)} & Swin-L & 23.6 \textcolor{teal}{(+9.4)} & 30.1 \textcolor{teal}{(+7.5)} & 36.0 \textcolor{teal}{(+9.7)} \\
        \bottomrule
    \end{tabular}
\end{adjustbox}
\caption{Performance comparison on the rare relation categories within the PSG dataset. DSGG consistently outperforms HiLo~\cite{zhou2023hilo}, demonstrating superior performance in rare relation categories.}
\label{table:psg_longtail_results}
\vspace{-5mm}
\end{table}

\paragraph{Model Parameters:}
Another important factor is that our model uses considerably fewer parameters when compared to~\cite{zhou2023hilo}, which incorporates two decoder streams: High-Low and Low-High. In particular, the DSGG model has a total parameter count of 44.2M, 107.1M, and 215.6M for the resnet-50, swin-b, and swin-l backbone networks, respectively. In contrast, the model proposed by ~\cite{zhou2023hilo} features 58.8M, 121.7M, and 230.3M parameters for the same backbone networks, respectively. The qualitative comparison of the Visual Genome and the PSG datasets is provided in the supplementary material.

\section{Conclusion}
\label{sec:conclusion}

In this paper, we introduce an innovative direct graph detection method for scene graph generation that simultaneously predicts objects and their relationships in an end-to-end fashion. Our approach employs novel graph-aware queries learned from dense scene graphs through relaxed sub-graph matching. Compositional tokens are utilized for learning embeddings for class, bounding-box, segmentation, and pairwise relations. Additionally, we incorporate relation distillation, re-scoring, and post-processing with logit adjustment for a unified end-to-end solution. Extensive experiments on scene graph generation (SGG) and panoptic scene graph generation (PSG) benchmark datasets demonstrate the superior performance of our method, surpassing state-of-the-art results significantly. Ablation studies assess the contribution of each model component, and we provide an analysis of our model's effectiveness in addressing challenges related to relational semantic overlap and long-tail issues. 

{
    \small
    \bibliographystyle{ieeenat_fullname}
    \bibliography{main}
}


\end{document}

%% file: preamble.tex
\usepackage[dvipsnames]{xcolor}

\usepackage{graphicx}
\usepackage{amsmath}
\usepackage{amssymb}
\usepackage{booktabs}
\usepackage{multirow}
\usepackage{bm}

\usepackage{comment}
\usepackage[utf8]{inputenc} 
\usepackage[T1]{fontenc}    
\usepackage{url}            
\usepackage{booktabs}       
\usepackage{amsfonts}       
\usepackage{nicefrac}       
\usepackage{microtype}      
\usepackage{xcolor}         
\usepackage{pifont}
\usepackage{bbm}
\usepackage{adjustbox}
\usepackage{graphicx, amsmath, amssymb, caption, subcaption, multirow, overpic, textpos}
\renewcommand{\paragraph}[1]{\vspace{1.25mm}\noindent\textbf{#1}}
\newlength\savewidth